\documentclass[10pt,twocolumn,letterpaper]{article}

\usepackage[utf8]{inputenc}
\usepackage{cvpr}
\usepackage{times}
\usepackage{epsfig}
\usepackage{graphicx}
\usepackage{amsmath,amsfonts,amssymb,amsthm}
\usepackage{color}
\usepackage{caption}
\usepackage{subcaption}
\usepackage{xspace}
\usepackage{array}
\usepackage{cite}
\usepackage{algorithm} %
\usepackage{algpseudocode} %
\usepackage{multirow}
\usepackage{booktabs}
\usepackage{soul}
\usepackage[pagebackref=true,breaklinks=true,letterpaper=true,colorlinks,bookmarks=false]{hyperref}

\newcommand{\bA}{\mathbf{A}}

\newcommand{\bD}{\mathbf{D}}

\newcommand{\bI}{\mathbf{I}}

\newcommand{\bK}{\mathbf{K}}

\newcommand{\bo}{\mathbf{o}}

\newcommand{\bR}{\mathbf{R}}
\newcommand{\bs}{\mathbf{s}}
\newcommand{\bt}{\mathbf{t}}

\newcommand{\bX}{\mathbf{X}}

\newcommand{\bz}{\mathbf{z}}

\newcommand{\bphi}{\boldsymbol{\phi}}

\newcommand{\nE}{\mathbb{E}}

\newcommand{\nR}{\mathbb{R}}

\newcommand{\cD}{\mathcal{D}}

\newcommand{\cL}{\mathcal{L}}

\newcommand{\cN}{\mathcal{N}}
\newcommand{\cO}{\mathcal{O}}

\newcommand{\figref}[1]{Fig.~\ref{#1}}

\newcommand{\tabref}[1]{Table~\ref{#1}}

\makeatletter
\DeclareRobustCommand\onedot{\futurelet\@let@token\@onedot}
\def\@onedot{\ifx\@let@token.\else.\null\fi\xspace}
\def\eg{e.g\onedot} 
\def\ie{i.e\onedot}

\def\wrt{wrt\onedot}

\def\etal{et~al\onedot}

\makeatother

\newcommand{\boldparagraph}[1]{\vspace{0.2cm}\noindent{\bf #1:} }
\newcommand{\boldparagraphwocolon}[1]{\vspace{0.2cm}\noindent{\bf #1} }

\definecolor{darkgreen}{rgb}{0,0.7,0}

\cvprfinalcopy
\def\cvprPaperID{5245}

\ifcvprfinal\fi

\begin{document}

\title{Towards Unsupervised Learning of Generative Models\\for 3D Controllable Image Synthesis}

\author{Yiyi Liao$^{1,2,*}$ \quad Katja Schwarz$^{1,2,*}$ \quad Lars Mescheder$^{1,2,3,^\dagger}$ \quad Andreas Geiger$^{1,2}$\\
\hspace{-0.35cm}$^1$Max Planck Institute for Intelligent Systems, Tübingen ~ $^2$University of Tübingen 
~ $^3$Amazon, Tübingen \\
{\tt\small {\{firstname.lastname\}}@tue.mpg.de}	
}

\maketitle
\renewcommand*{\thefootnote}{\fnsymbol{footnote}}
\footnotetext{$^*$ Joint first authors with equal contribution.} %
\footnotetext{$^\dagger$ This work was done prior to joining Amazon.} 
\renewcommand*{\thefootnote}{\arabic{footnote}}

\begin{abstract}
In recent years, Generative Adversarial Networks have achieved impressive results in photorealistic image synthesis. This progress nurtures hopes that one day the classical rendering pipeline can be replaced by efficient models that are learned directly from images. However, current image synthesis models operate in the 2D domain where disentangling 3D properties such as camera viewpoint or object pose is challenging. Furthermore, they lack an interpretable and controllable representation. Our key hypothesis is that the image generation process should be modeled in 3D space as the physical world surrounding us is intrinsically three-dimensional. We define the new task of 3D controllable image synthesis and propose an approach for solving it by reasoning both in 3D space and in the 2D image domain. We demonstrate that our model is able to disentangle latent 3D factors of simple multi-object scenes in an unsupervised fashion from raw images. Compared to pure 2D baselines, it allows for synthesizing scenes that are consistent wrt. changes in viewpoint or object pose. We further evaluate various 3D representations in terms of their usefulness for this challenging task.
\end{abstract}

\begin{figure}[t!]
	\center
	\includegraphics[width=\linewidth]{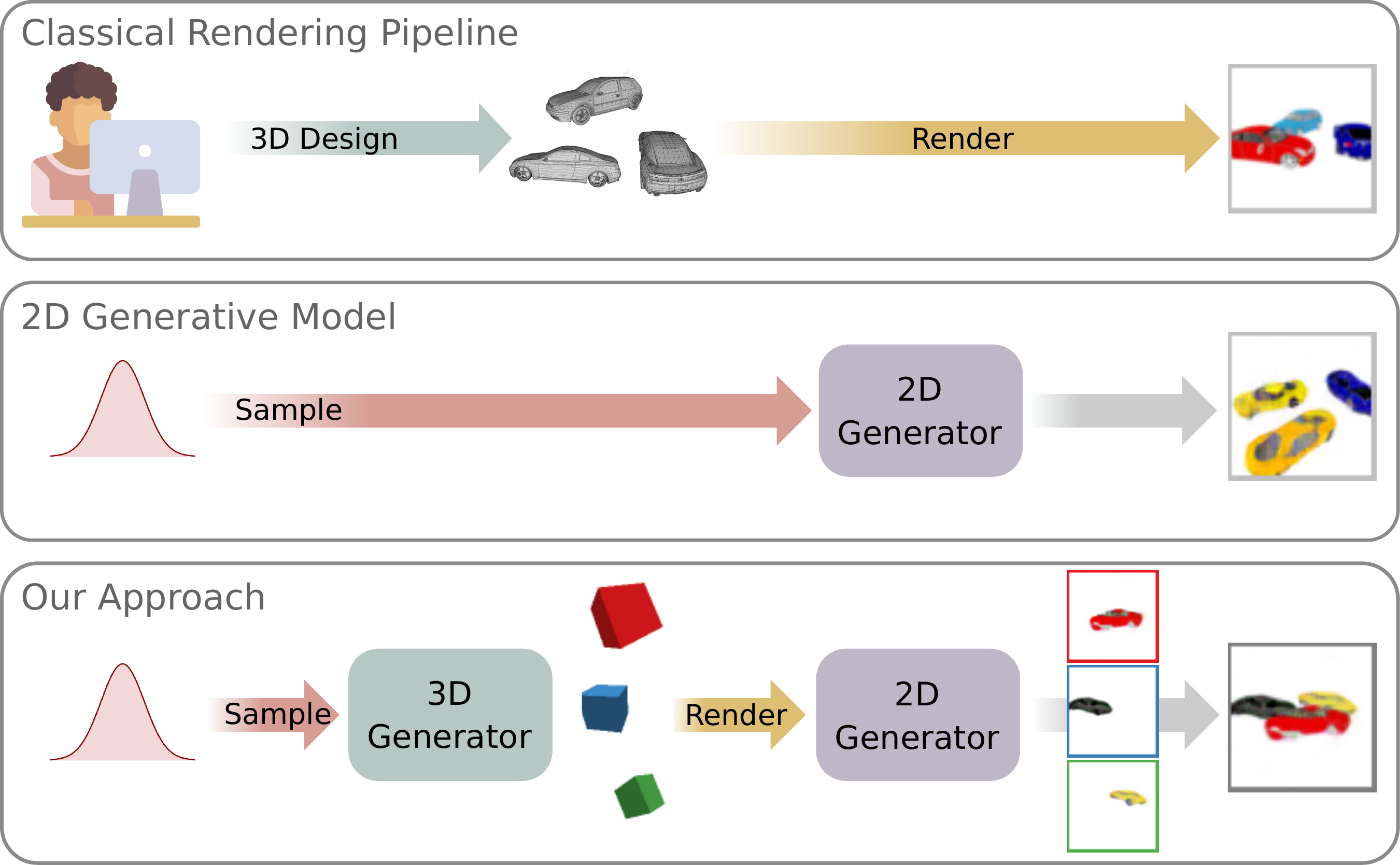}
	\caption{{\bf Motivation.}
	While classical rendering algorithms require a holistic scene representation, image-based generative neural networks are able to synthesize images from a latent code and can hence be trained directly from images.
	We propose to consider the generative process in both 3D and 2D space,
	yielding a 3D controllable image synthesis model which learns both the 3D content creation process as well as the rendering process jointly from raw images.
	}
	\label{fig:teaser}
	\vspace{-0.3cm}
\end{figure}

\section{Introduction}

Realistic image synthesis is a fundamental task in computer vision and graphics with many applications including gaming, simulation, virtual reality and data augmentation.
In all these applications, it is essential that the image synthesis algorithm allows for controlling the 3D content and produces coherent images when changing the camera position and object locations.
Imagine exploring a virtual reality room. When walking around and manipulating objects in the room, it is essential to observe coherent images, \ie manipulating the pose of one object should not alter properties (such as the color) of any other object in the scene.

Currently, image synthesis in these applications is realized using rendering engines (\eg, OpenGL) as illustrated in \figref{fig:teaser} (top).
This approach provides full control over the 3D content since the input to the rendering algorithm is a holistic description of the 3D scene.
However, creating photorealistic content and 3D assets for a movie or video game is an extremely expensive and time-consuming process and requires the concerted effort of many 3D artists.
In this paper, we therefore ask the following question:

\textit{Is it possible to learn the simulation pipeline including 3D content creation from raw 2D image observations?}

Recently, Generative Adversarial Networks (GANs) have achieved impressive results for photorealistic image synthesis~\cite{Mescheder2018ICML,Karras2018ICLR,Karras2019CVPR} and hence emerged as a promising alternative to classical rendering algorithms,
see \figref{fig:teaser} (middle).
However, a major drawback of image-based GANs is that the learned latent representation is typically ``3D entangled''.
That is, the latent dimensions do not automatically expose physically meaningful 3D properties such as camera viewpoint, object pose or object entities.
As a result, 2D GANs fall short in terms of controllability compared to classical rendering algorithms.
This limits their utility for many applications including virtual reality, data augmentation and simulation.

\begin{figure}
	\center
	\begin{minipage}[t]{0.0465\linewidth}
		\includegraphics[width=\linewidth]{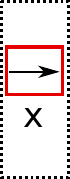}
	\end{minipage}\begin{minipage}[t]{0.953\linewidth}
		    \includegraphics[width=\linewidth]{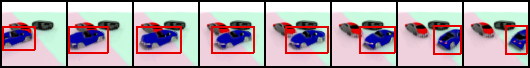}
	\end{minipage}
	\begin{minipage}[t]{0.0465\linewidth}
		\includegraphics[width=\linewidth]{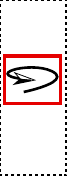}
	\end{minipage}\begin{minipage}[t]{0.953\linewidth}
		    \includegraphics[width=\linewidth]{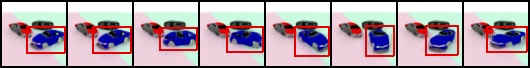}
	\end{minipage}
	\begin{minipage}[t]{0.0465\linewidth}
		\includegraphics[width=\linewidth]{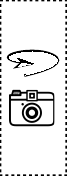}
	\end{minipage}\begin{minipage}[t]{0.953\linewidth}
		    \includegraphics[width=\linewidth]{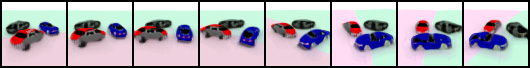}
	\end{minipage}
	\begin{minipage}[t]{0.0465\linewidth}
		\includegraphics[width=\linewidth]{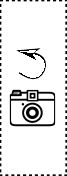}
	\end{minipage}\begin{minipage}[t]{0.953\linewidth}
		    \includegraphics[width=\linewidth]{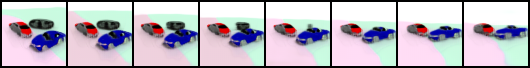}
	\end{minipage}
    \caption{{\bf 3D Controllable Image Synthesis.} We propose a model for 3D controllable image synthesis which allows for manipulating the viewpoint and 3D pose of individual objects.
    Here, we illustrate consistent translation in 3D space, object rotation and viewpoint changes.
    } 
    \label{fig:controllability}
    \vspace{-0.5cm}
\end{figure}%

\boldparagraph{Contribution}
In this work, we propose the new task of \textit{3D Controllable Image Synthesis}.
We define this task as an unsupervised learning problem, where a 3D controllable generative image synthesis model that allows for manipulating 3D scene properties is learned without 3D supervision.
The 3D controllable properties include the 3D pose, shape and appearance of \textit{multiple} objects as well as the viewpoint of the camera.
Learning such a model from 2D supervision alone is very challenging as the model must reason about the modular composition of the scene as well as physical 3D properties of the world such as light transport.

Towards a solution to this problem, we propose a novel formulation that combines the advantages of deep generative models with those of traditional rendering pipelines.
Our approach enables 3D controllability when synthesizing new content while learning meaningful representations from images alone.
Our key idea is to learn the image generation process jointly in 3D and 2D space by combining a 3D generator with a differentiable renderer and a 2D image synthesis model as illustrated in \figref{fig:teaser} (bottom).
This allows our model to learn abstract 3D representations which conform to the physical image formation process, thereby retaining interpretability and controllability.
We demonstrate that our model is able to disentangle the latent 3D factors of simple scenes composed of multiple objects and allows for synthesizing images that are consistent \wrt changes in viewpoint or object pose as illustrated in \figref{fig:controllability}.
Our code and data are provided at \url{https://github.com/autonomousvision/controllable_image_synthesis}.

\section{Related Work}
\boldparagraph{Image Decomposition}
Several works have considered the problem of decomposing a scene, conditioned on an input image~\cite{Burgess2019ARXIV,Eslami2016NIPS,Greff2016NIPS,Greff2017NIPS,Greff2019ICML,Kosiorek2018NIPS,Stanic2019NIPSWorkshop,Steenkiste2018ICLR}.
One line of work uses RNNs to sequentially decompose the scene~\cite{Burgess2019ARXIV,Eslami2016NIPS,Kosiorek2018NIPS,Stanic2019NIPSWorkshop}, whereas other approaches iteratively refine the image segmentation~\cite{Greff2016NIPS,Greff2017NIPS,Steenkiste2018ICLR,Greff2019ICML}.
While early works consider very simple scenes~\cite{Eslami2016NIPS,Kosiorek2018NIPS} or binarized images~\cite{Greff2016NIPS,Greff2017NIPS,Steenkiste2018ICLR}, recent works~\cite{Burgess2019ARXIV,Greff2019ICML} also handle scenes with occlusions.

In contrast to the problem of learning controllable generative models, the aforementioned approaches are purely discriminative and are not able to synthesize novel scenes. Furthermore, they operate purely in the 2D image domain and thus lack an understanding of the physical 3D world.

\boldparagraph{Image Synthesis}
Unconditional Generative Adversarial Networks (GANs)~\cite{Radford2015ARXIV,Mescheder2017NIPS,Goodfellow2014NIPS} have greatly advanced photo-realistic image synthesis by learning deep networks that are able to sample from the natural image manifold.
More recent works condition these models on additional input information (\eg, semantic segmentations) to guide the image generation process~\cite{Ashual2019ICCV,Chen2017ICCV, Isola2017CVPR, Johnson2018CVPR, li2019layoutgan, Reed2016NIPS, Turkoglu2019AAAI, Wang2018CVPR, Zhao2019CVPR}.

Another line of work learns interpretable, disentangled features directly from the input data for manipulating the generated images~\cite{Chen2016NIPS,Reed2014ICML,Karras2019CVPR,Zhao2018ECCV}.
More recent work aims at controlling the image generation process at the object level~\cite{Shaham2019ICCV,Yang2017ICLR,Engelcke2019ARXIV}.
The key insight is that this splits the complex task of image generation into easier subproblems which benefits the quality of the generated images.

However, all of these methods operate based on a 2D understanding of the scene. 
While some of these works are able to disentangle 3D pose in the latent
representation~\cite{Chen2016NIPS}, full 3D control in terms of object
translations and rotations as well as novel viewpoints remains a challenging task.
In this work, we investigate the problem of learning an interpretable intermediate representation at the object-level that is able to provide \textit{full 3D control} over all objects in the scene. We also demonstrate that reasoning in 3D improves consistency wrt. pose and viewpoint changes.

\boldparagraph{3D-Aware Image Synthesis}
Several works focus on 3D controllable image synthesis with 3D supervision~\cite{Wang2016ECCV,Zhu2018NIPS} or using 3D information as input~\cite{Alhaija2018ACCV}.
In this work, we instead focus on learning from 2D images alone.
The problem of learning discriminative 3D models
from 2D images is often addressed using differentiable rendering~\cite{Henzler2019ICCV, Kato2018CVPR, Liu2019soft, Navaneet2019CVPRW}. 
Our work is related to these works in that we also exploit a differentiable renderer.
However, unlike the aforementioned methods, we combine differentiable rendering with generative models in 3D and 2D space.
We learn an abstract 3D representation that can be transformed into photo-realistic images via a neural network.
Previous works in this direction can be categorized as either implicit or explicit, depending on whether the learned features have a physical meaning.

Implicit methods~\cite{Worrall2017ICCV,Rhodin_2018_ECCV} apply rotation to a 3D latent feature vector to generate transformed images through a multilayer perceptron.
This feature vector is a global representation as it influences the entire image.

In contrast, explicit methods~\cite{Henzler2019ICCV,nguyen2019hologan, sitzmann2019deepvoxels} exploit feature volumes that can be differentiably projected into 2D image space.
DeepVoxels~\cite{sitzmann2019deepvoxels} proposes a novel view synthesis approach which unprojects multiple images into a 3D volume and subsequently generates new images by projecting the 3D feature volume to novel viewpoints.
Albeit generating high quality images, their model is not generative and requires hundreds of images from the same object for training. 
In contrast, HoloGAN~\cite{nguyen2019hologan} and {\sc platonic}GAN~\cite{Henzler2019ICCV} are generative models that are able to synthesize 3D consistent images of single objects by generating volumetric 3D representations based on a differentiable mapping.

However, all these approaches
are only able to learn models for single objects or static scenes.
Therefore, their controllability is limited to object-centric rotations or camera viewpoint changes.
In contrast, we propose a model for learning disentangled 3D representations which allows for manipulating multiple objects in the scene individually.

\begin{figure*}[h!]
	\includegraphics[width=\linewidth]{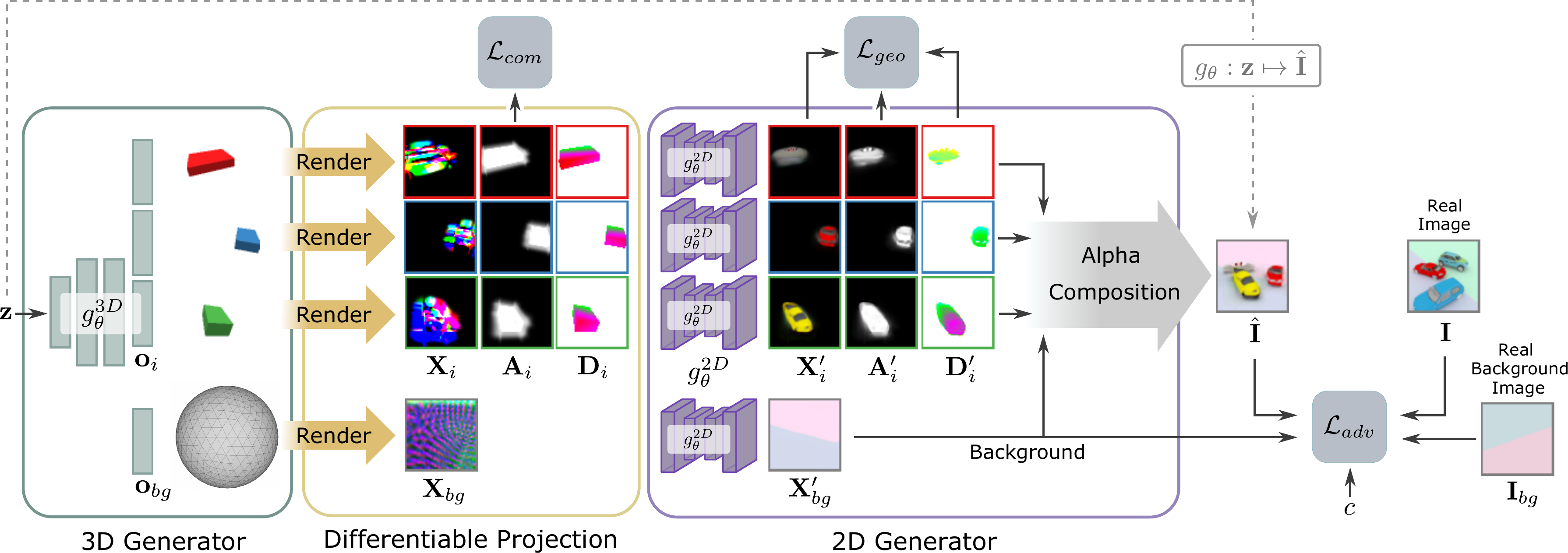}
	\caption{{\bf Method Overview.} Our model comprises three main parts: A \textit{3D Generator} which maps a latent code $\bz$ drawn from a Gaussian distribution into a set of abstract 3D primitives $\{\bo_i\}$, a \textit{Differentiable Projection} layer which takes each 3D primitive $\bo_i$ as input and outputs a feature map $\bX_i$, an alpha map $\bA_i$ and a depth map $\bD_i$, and a \textit{2D Generator} which refines them and produces the final composite image $\hat{\bI}$. We refer to the entire model as $g_\theta:\bz \mapsto \hat{\bI}$ and train it by minimizing a compactness loss $\cL_{com}$, a geometric consistency loss $\cL_{geo}$ and an adversarial loss $\cL_{adv}$ which compares the predicted image $\hat{\bI}$ to images $\bI$ from the training set. The last primitive $\bo_{bg}$ generates the image background $\bX'_{bg}$. The flag $c$ determines if the adversarial loss compares full composite images ($c=1$) or only the background ($c=0$) to the training set.
	}
	\label{fig:overview}
	\vspace{-0.5cm}
\end{figure*}

\section{Method}\label{sec:method}

Our goal is to learn a controllable generative model for image synthesis from raw image data.
We advocate that image synthesis should benefit from an interpretable 3D representation as this allows to explicitly take into account the image formation process (\ie, perspective projection, occlusion, light transport), rather than trying to directly learn the distribution of 2D images. We thus model the image synthesis process jointly in 3D and 2D space by first generating an abstract 3D representation that is subsequently projected to and refined in the 2D image domain.

Our model is illustrated in \figref{fig:overview}.
We first sample a latent code that represents all properties of the generated scene.
The latent code is passed to a 3D generative model which generates a set of 3D abstract object representations in the form of 3D primitives as well as a background representation.
The 3D representations are projected onto the image plane where a 2D generator transforms them into object appearances and composites them into a coherent image.
As supervision at the primitive level is hard to acquire, we propose an approach based on adversarial training which does not require 3D supervision.

We now specify our model components in detail.
One contribution of our paper is to analyze and compare various 3D representations regarding their suitability for this task.
We thus first define the 3D primitive representations that we consider in this work.
In the next section, we describe our model for generating the primitive parameters as well as the rendering step that projects the 3D primitives to 2D feature maps in the image plane.
Finally, we describe the 2D generator which synthesizes and composites the image as well as the loss functions that we use for training our model in an end-to-end fashion from unlabeled images.

\subsection{3D Representations}

We aim for a disentangled 3D object representation that is suitable for end-to-end learning while at the same time allowing full control over the geometric properties of each individual scene element, including the object's scale and pose. We refer to this abstract object representation using the term ``primitive'' in the following.

Further, we use the term ``disentangled representation'' to describe a set of distinct, informative factors of variations in the observations~\cite{Bengio2013PAMI,Locatello2019ICML}. We model these factors with a set of 3D primitives $\cO=\{\bo_{bg},\bo_1, \dots, \bo_N\}$ to disentangle the foreground objects and the scene background. Each \textit{foreground object} is described by a set of attributes $\bo_i = (\bs_i,\bR_i,\bt_i,\bphi_i)$ 
which represent the properties of each object instance. Here,
$\bs_i\in\nR^3$ denotes the 3D object scale, $\bR_i\in SO(3)$ and $\bt_i\in \nR^3$ are the primitive's 3D pose parameters, and $\bphi_i$ is a feature vector determining its appearance. To model the \textit{scene background}, we introduce an additional background primitive $\bo_{bg}$ defined accordingly.

As it is unclear which 3D representation is best suited for our task, we compare and discuss the following feature representations in our experimental evaluation:

\boldparagraph{Point Clouds}
Point clouds are one of the simplest forms to represent 3D information.
We represent each primitive with a set of sparse 3D points.
More specifically, each $\bphi_i = (\bphi_i^l,\bphi_i^f)$ represents the location $\bphi_i^l \in \nR^{M\times 3}$ and feature vectors $\bphi_i^f \in \nR^{M \times K}$ of $M$ points with latent feature dimension $K$.
We apply scaling $\bs_i$, rotation $\bR_i$ and translation $\bt_i$ to the point location $\bphi_i^l$.

\boldparagraph{Cuboids and Spheres}
We further consider cuboids and spheres as representations which are more closely aligned with classical mesh-based models used in computer graphics.
In this case, the feature vector $\bphi_i$ represents a texture map that is attached to the surface of the cuboid or sphere, respectively.
The geometry is determined by scaling, rotating and translating the primitive via $\bs_i$, $\bR_i$ and $\bt_i$.

\boldparagraph{Background}
In order to allow for changes in the camera viewpoint, we do not use a vanilla 2D GAN for generating a background image.
Instead, we represent the background using a spherical environment map.
More specifically, we attach the background feature map $\bphi_{bg}$ as a texture map to the inside of a sphere located at the origin, \ie, $\bR_{bg}=\bI$ and $\bt_{bg}=\mathbf{0}$.
The scale $\bs_{bg}$ of the background sphere is fixed to a value that is large enough to inclose the entire scene including the camera.

\subsection{3D Generator}
\label{sec:3d_generator}

The first part of our neural network is an implicit generative 3D model which generates the set of latent 3D primitives $\{\bo_{bg},\bo_1, \dots, \bo_N\}$ from a noise vector $\bz \sim \cN(\mathbf{0},\bI)$.
We use a fully connected model architecture with a fixed number of $N+1$ heads to generate the attributes for the $N$ foreground primitives and the background. Note, that early layers share weights such that all primitives are generated jointly and share information.
More formally we have:
\begin{equation}
g^{3D}_\theta: \bz \mapsto \{\bo_{bg},\bo_1, \dots, \bo_N\}
\end{equation}
where $\theta$ denote the parameters of the 3D generation layer.
\figref{fig:overview} (left) illustrates our 3D generator. Details about the network architecture can be found in the supplementary.

\subsection{Differentiable Projection}

The differentiable projection layer takes each of the generated primitives $\bo_i \in \cO$ as input and converts each of them separately into a feature map $\bX_i\in\nR^{W \times H \times F}$ where $W \times H$ are the dimensions of the output image and $F$ denotes the number of feature channels.
In addition, it generates a coarse alpha map $\bA_i\in\nR^{W \times H}$ and an initial depth map $\bD_i\in\nR^{W \times H}$ for each primitive which are further refined by the 2D generator that is described in the next section. We implement the differentiable projection layer as follows, assuming 
(without loss of generality)
a fixed calibrated camera with intrinsics $\bK\in\nR^{3\times3}$.

\boldparagraph{Cuboids, Spheres and Background}
As cuboids, spheres and the background are represented in terms of a surface mesh, we exploit a differentiable mesh renderer to project them onto the image domain.
More specifically, we use the Soft Rasterizer of Liu \etal~\cite{Liu2019soft} and adapt it to our purpose.
In addition to the projected features $\bX_i$, we obtain an alpha map $\bA_i$ by smoothing the silhouette of the projected object using a Gaussian kernel.
We further output the depth map $\bD_i$ of the projected mesh.

\boldparagraph{Point Clouds}
For point clouds we follow prior works~\cite{Insafutdinov2018NIPS} and represent them in a smooth fashion using isotropic Gaussians.
More specifically, we project all features onto an initial empty feature map $\bX_i$ and smooth the result using a Gaussian blur kernel.
This allows us to learn both the locations $\bphi_i^l$ as well as the features $\bphi_i^f$ while also back-propagating gradients to the pose parameters $\{\bs_i,\bR_i,\bt_i\}$ of the $i$'th primitive.
Since point clouds are sparse, we determine the initial alpha map $\bA_i$ 
and depth map $\bD_i$ by additionally projecting a cuboid with the same scale, rotation and translation onto the image plane as described above.

\subsection{2D Generator}
\label{sec:2d_generator}

Learning a 3D primitive for each object avoids explicitly reconstructing their exact 3D models, while the projected features are abstract.
We learn to transform this abstract representation into a photorealistic image using a 2D generative model.
More specifically, for each primitive, we use a fully convolutional network which takes the features, the initial alpha map and the initial depth map as input and refines it, yielding a color rendering, a refined alpha map and a refined depth map of the corresponding object
\begin{equation}
g^{2D}_\theta: \bX_i,\bA_i,\bD_i \mapsto \bX'_i,\bA'_i,\bD'_i
\end{equation}
where $\theta$ denote the parameters of the network that are shared across the objects and the background\footnote{We slightly abuse notation and use the same symbol $\theta$ to describe the parameters of the 2D and the 3D generative models}.
We use a standard encoder-decoder structure based on ResNet~\cite{He2016CVPR} as our 2D generator. See supplementary material for details.

It is important to note that we perform amodal object prediction, \ie, all primitives are predicted separately without taking into account occlusion.
Thus, our 2D generator can learn to generate the entire object for each primitive, even if it is partially occluded in some of the images.

The final step in our 2D generation layer is to combine the individual predictions into a composite image $\hat{\bI}$.
Towards this goal, we fuse all predictions using alpha composition in ascending order of depth at each pixel.
Implementation details are provided in the supplementary.

\subsection{Loss Functions}

We train the entire model end-to-end using adversarial training. 
Importantly, we do not rely on supervision in the form of labeled 3D primitives, instance segmentations or pose annotations. The only input to our method are images of
scenes with various number of objects in 
different poses, from varying viewpoints and with varying backgrounds.

Learning such a model without supervision is a challenging task with many ambiguities. The model could for instance learn to explain two objects with a single primitive or even to generate all foreground objects with the background model, setting all alpha maps $\bA'_i$ to zero. We therefore introduce multiple loss functions that encourage a disentangled and interpretable 3D representation while at the same time synthesizing images from the training data distribution. Our loss $\cL$ is composed of three terms:
\begin{equation}
\cL(\theta,\psi)  =  \sum_{c\in\{0,1\}} \cL_{adv}(\theta,\psi,c) + \cL_{com}(\theta) + \cL_{geo}(\theta)
\end{equation}

\boldparagraph{Adversarial Loss}
We use a standard adversarial loss~\cite{Goodfellow2014NIPS} to encourage that the images generated by our model follow the data distribution. Let $\bI$ denote an image sampled from the data distribution $p_{\cD}(\bI)$ and let $g_\theta : \bz \mapsto \hat{\bI}$ denote the entire generative model.
Let further $d_\psi(\bI)$ denote the discriminator. Our adversarial loss is formulated as follows
\begin{equation}
\begin{split}
\cL_{adv}(\theta, \psi,c) & = \nE_{p(\bz)} [f(d_\psi(g_\theta(\bz,c), c)) ] \\
& + \nE_{p_{\cD}(\bI|c)} [f(-d_\psi(\bI, c))]
\end{split}
\end{equation}
with $f(t)=-\log(1+\exp(-t))$.
Note that we condition both the generator $g_\theta(\bz,c)$ as well as the discriminator $d_\psi(\bI, c)$ on an additional variable $c\in\{0,1\}$ which determines if the generated/observed image is a full composite image ($c=1$) or a background image ($c=0$).
This condition is helpful to disentangle foreground objects from the background as evidenced by our experiments.
For training our model, we thus collect two datasets: one that includes foreground objects and one with empty background scenes.

\boldparagraph{Compactness Loss}
To bias solutions towards compact representations and to encourage the 3D primitives to tightly encase the objects, 
we minimize the projected shape of each object. We formulate this constraint as a penalty on the $l_1$-norm of 
the alpha maps $\bA_i$ of each object:
\begin{equation}
\cL_{com}(\theta) = \nE_{p(\bz)} \left[ \sum_{i=1}^N\max\left(\tau, \frac{{\Vert \bA_i \Vert}_{1}}{H\times W} \right) \right]
\end{equation}
Here, $\tau=0.1$ is a truncation threshold which avoids shrinkage below a fixed minimum size and $\bA_i$ depends\footnote{We omit this dependency here for clarity.} on the model parameters $\theta$ and the latent code $\bz$.

\boldparagraph{Geometric Consistency Loss}
To favor solutions that are consistent across camera viewpoints and 3D object poses, we follow~\cite{Noguchi2019ARXIV} and encourage the learned generative model to conform to multi-view geometry constraints.
For instance, a change in the pose parameters $(\bR_i,\bt_i)$ should change the pose of the object but it should not alter neither its color nor its identity.
We formulate this constraint as follows:
\begin{equation}
\begin{split}
\cL_{geo}(\theta) &= \nE_{p(\bz)} \left[\sum_{i=1}^N {\Vert \bA'_i \odot (\bX'_i - \tilde{\bX}'_i)\Vert}_1 \right]\\
& + \nE_{p(\bz)}\left[ \sum_{i=1}^N{\Vert\bA'_i \odot (\bD'_i - \tilde{\bD}'_i)\Vert}_1\right]
\end{split}
\label{eq:loss_geo}
\end{equation}
Here, $(\bX'_i,\bD'_i)$ are the outputs of the 2D generator for latent code $\bz$. $(\tilde{\bX}'_i,\tilde{\bD}'_i)$ correspond to the outputs when using the same latent code $\bz$, but adding random noise to the pose parameters of each primitive and warping the results back to the original viewpoint.
The warping function is determined by the predicted depth and the relative transformation between the two views.
The operator $\odot$ denotes elementwise multiplication
and makes sure that geometric consistency is only enforced inside the foreground region, \ie, where $\bA'_i = 1$.
Intuitively, the geometric consistency loss encourages objects with the same appearance but which are observed from a different view to agree both in terms of appearance and depth.
At the same time it serves as an unsupervised loss for training the depth prediction model.
For details on the warping process, we refer the reader to the supplemental material.

\subsection{Training}

We train our model using RMSprop~\cite{tieleman2012lecture} and a learning rate of $10^{-4}$. To stabilize GAN training, we use gradient penalties~\cite{Mescheder2018ICML} and spectral normalization~\cite{Miyato2018ICLR} for the discriminator. 
Following~\cite{Karras2019CVPR} we use adaptive instance normalization~\cite{Huang2017ICCV} for the 2D generator. We randomly sample the camera viewpoint during training from an upper-half sphere around the origin.

\section{Experiments}\label{sec:results}

\begin{figure}
	\begin{minipage}{0.25\linewidth}
		\includegraphics[width=0.5\linewidth]{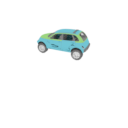}%
		\includegraphics[width=0.5\linewidth]{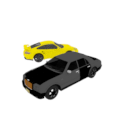}\\
		\includegraphics[width=0.5\linewidth]{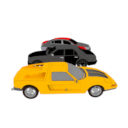}%
		\includegraphics[width=0.5\linewidth]{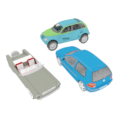}\\
		\centering Car w/o BG
	\end{minipage}%
	\begin{minipage}{0.25\linewidth}
		\includegraphics[width=0.5\linewidth]{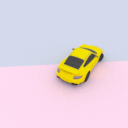}%
		\includegraphics[width=0.5\linewidth]{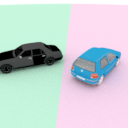}\\
		\includegraphics[width=0.5\linewidth]{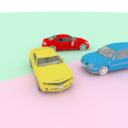}%
		\includegraphics[width=0.5\linewidth]{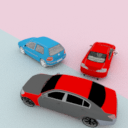}\\
		\centering Car with BG
	\end{minipage}~%
	\begin{minipage}{0.25\linewidth}
		\includegraphics[width=0.5\linewidth]{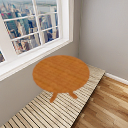}%
		\includegraphics[width=0.5\linewidth]{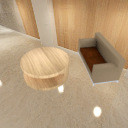}\\
		\includegraphics[width=0.5\linewidth]{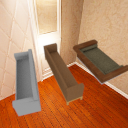}%
		\includegraphics[width=0.5\linewidth]{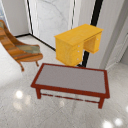}\\
		\centering Indoor
	\end{minipage}~%
	\begin{minipage}{0.25\linewidth}
		\includegraphics[width=0.5\linewidth]{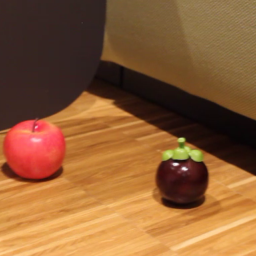}%
		\includegraphics[width=0.5\linewidth]{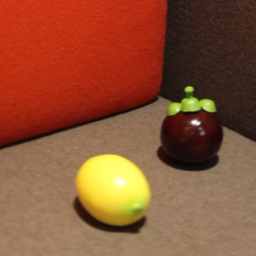}\\
		\includegraphics[width=0.5\linewidth]{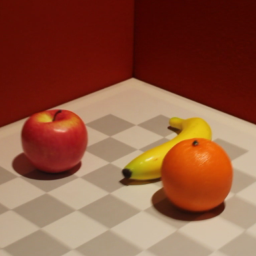}%
		\includegraphics[width=0.5\linewidth]{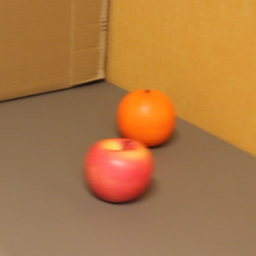}\\
		\centering Fruit
	\end{minipage}
	\caption{{\bf Datasets.} Random samples from each dataset.} 
	\label{fig:dataset_samples}
	\vspace{-0.5cm}
\end{figure}

In this section, we first compare our approach to several baselines on the task of 3D controllable image generation, both on synthetic and real data.
Next, we conduct a thorough ablation study to better understand the influence of different representations and architecture components.

\boldparagraph{Dataset}
We render synthetic datasets using objects from  ShapeNet~\cite{Chang2015ARXIV}, considering three datasets with varying difficulty.
Two datasets contain cars, one with and the other without background.
For both datasets, we randomly sample $1$ to $3$ cars from a total of $10$ different car models. 
Our third dataset is the most challenging of these three. It comprises indoor scenes containing objects of different categories, including chairs, tables and sofas. As background we use empty room images from Structured3D~\cite{Zheng2019ARXIV}, a synthetic dataset with photo-realistic 2D images.
For each dataset we render 48k real images~$\bI$ and 32k unpaired background images $\bI_{bg}$ for training, as well as 9.6k images for validation and testing.
The image resolution is $64\times 64$ for all datasets.
In addition to the synthetic datasets, we apply our method to a real world dataset containing $800$ images of $5$ fruits with $5$ different backgrounds. 
In order to train our model, we also collect a set of unpaired background images.
Sample images from our datasets are shown in \figref{fig:dataset_samples}.

\boldparagraph{Baselines}
We first compare our method to several baselines on the car dataset and the indoor dataset: 
\textbf{Vanilla GAN}~\cite{Mescheder2018ICML}: a state-of-the-art 2D GAN with gradient penalties.
\textbf{Layout2Im}~\cite{Zhao2019CVPR}: a generative 2D model that generates images conditioned on 2D bounding boxes.
\textbf{2D baseline}: To investigate the advantage of our 3D representation, we implement a 2D version of our method by replacing the 3D generator by a 2D feature generator, i.e. learning a set of 2D feature maps instead of 3D primitives. 
\textbf{Ours w/o $c$}: We train our method without background images $\bI_{bg}$ to investigate if the method is able to disentangle foreground and background without extra supervision.
All implementation details are provided in the supplementary.

\boldparagraph{Metrics}
We measure the quality of the generated images using the Fréchet Inception Distance (FID)~\cite{Heusel2017NIPS}. More specifically, we compute the FID score between the distribution of real images $\bI$ and the distribution of generated images $\hat{\bI}$, 
excluding the background images $\bI_{bg}$. 
In our ablation study, we also report FID$_i$ 
to quantify the disentanglement of object instances.
FID$_i$  measures the distance between rendered single-object images and our images for each primitive before alpha composition.
Besides evaluating on the generated samples, we also apply additional random rotation and translation to the 3D primitives and report FID scores on these transformed samples (FID$_\bt$, FID$_\bR$). Similarly, we apply random translation to Layout2Im and the 2D baseline. 
We investigate how well our model captures the underlying 3D distribution in the supplementary.

\subsection{Controllable Image Generation}

We now report our results for the challenging task of 3D controllable image generation on synthetic and real data.

\boldparagraph{Synthetic Data}
\tabref{tab:baselines} compares the FID scores on the Car and Indoor datasets. 
Comparing the quantitative results,
we see that our method can achieve competitive FID score compared to Layout2Im. However, Layout2Im requires 2D bounding boxes as supervision while our method operates in unsupervised manner. 

To test our hypothesis if 3D representations are helpful for controllable image synthesis, we measure FID scores FID$_\bt$ and FID$_\bR$ on transformed samples.
Our results show that the FID scores are relatively stable \wrt random rotations and translations, demonstrating that the perturbed images follow the same distribution.
Without background supervision (w/o $c$), our method can disentangle foreground from background if the background appearance is simple (\eg, Car dataset).
However, in the case of more complex background appearance (\eg, Indoor dataset) the foreground primitives vanish while the background generates the entire image, resulting in a higher FID score.
In contrast, with unpaired background images as supervision, our method is able to disentangle foreground from background even in the presence of complex backgrounds.

\begin{figure}
  \begin{center}
  \setlength\tabcolsep{0.2em}
  \begin{tabular}{m{0.5em}m{22.5em}}
    {\rotatebox[origin=c]{90}{\small{\cite{Zhao2019CVPR}  ($\bt$)}}} &
    \includegraphics[width=\linewidth]{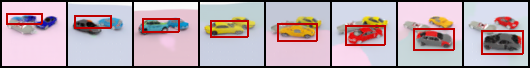}\\
    {\rotatebox[origin=c]{90}{\small{2D  ($\bt$)}}} &
    \includegraphics[width=\linewidth]{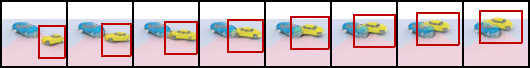} \\
    {\rotatebox[origin=c]{90}{\small{Ours ($\bt$)}}} &
    \includegraphics[width=\linewidth]{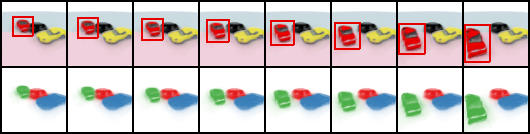} \\
    {\rotatebox[origin=c]{90}{\small{Ours ($\bR$)}}} &
    \includegraphics[width=\linewidth]{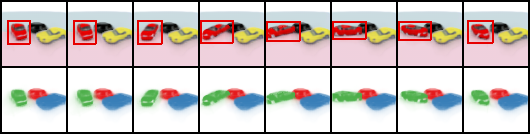}
  \end{tabular}
  \end{center}
  \vspace{-0.5cm}
  \caption{{\bf Car Dataset.}  
	We translate one object for all methods. Additionally, we rotate one object with our method which cannot be achieved with the baselines.  
  }
  \label{fig:translation_car}
\end{figure}

\begin{figure}
  \begin{center}
  \setlength\tabcolsep{0.2em}
  \begin{tabular}{m{0.5em}m{22.5em}}
    {\rotatebox[origin=c]{90}{\small{\cite{Zhao2019CVPR} ($\bt$)}}} &
    \includegraphics[width=\linewidth]{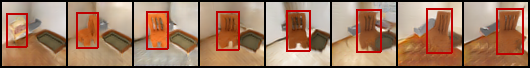}\\
    {\rotatebox[origin=c]{90}{\small{2D ($\bt$)}}} &
    \includegraphics[width=\linewidth]{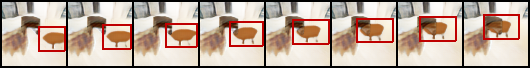}\\
    {\rotatebox[origin=c]{90}{\small{Ours ($\bt$)}}} &
    \includegraphics[width=\linewidth]{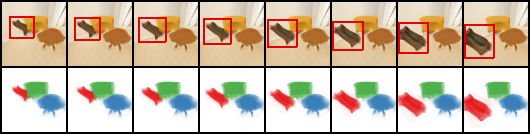}\\
    {\rotatebox[origin=c]{90}{\small{Ours ($\bR$)}}} &
    \includegraphics[width=\linewidth]{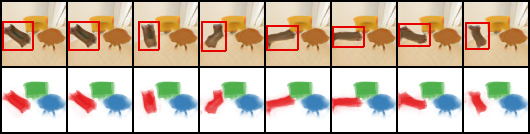}\\
  \end{tabular}
  \end{center}
  \vspace{-0.5cm}
  \caption{{\bf Indoor Dataset.} 
We translate one object for all methods. Additionally, we rotate one object with our method which cannot be achieved with the baselines.   
  } 
  \label{fig:translation_indoor}
  \vspace{-0.1cm}
\end{figure}

In \figref{fig:translation_car} and \figref{fig:translation_indoor}, we show qualitative results when transforming the objects. 
By translating the 2D bounding boxes, Layout2Im achieves 2D controllability that can even handle occlusion. However, the images lack consistency: moving an object changes its identity, indicating a failure of the model to disentangle the latent factors.
Moreover, Layout2Im fuses the objects with a recurrent module. Therefore, manipulating one object also affects other objects in the scene as well as the background. Note for example how the black car in the leftmost image in \figref{fig:translation_car} becomes yellow during translation or how the background in both \figref{fig:translation_car} and \figref{fig:translation_indoor} varies.
Our 2D baseline is able to better disentangle objects, while it struggles to learn the correct occlusion relationship.
In contrast, our model is able to correctly disentangle the latent factors and produces images that are consistent \wrt object rotation and translation.

\begin{table}[t]
\begin{center}
\begin{small}
\begin{tabular}{@{}lc@{\hskip 5pt}c@{\hskip 5pt}cc@{\hskip 5pt}c@{\hskip 5pt}c@{}}
\toprule
& \multicolumn{3}{c}{Car} & \multicolumn{3}{c}{Indoor} \\ \cmidrule(l){2-4} \cmidrule(l){5-7} 
{} &         FID &         FID$_\bt$  &           FID$_\bR$ &         FID &         FID$_\bt$  &           FID$_\bR$\\ 
\hline
Vanilla GAN~\cite{Mescheder2018ICML} 	& 	\textbf{43} 	&	 -- 	& -- 	& 	89 & -- & -- \\
Layout2Im~\cite{Zhao2019CVPR} 	& \textbf{43} &56 & --&\textbf{84} &93 & -- \\
2D Baseline		& 	80 &	79 		&  -- 	&	107 &         102 &  --  \\

Ours (w/o $c$) &     65 &   71 	&  75 &   120 &    120 &         120 \\
Ours &  44 &          \textbf{54} &          \textbf{66}  &  88 &          \textbf{90} &         \textbf{100} \\
\bottomrule
\end{tabular}

\end{small}
\caption{{{\bf FID on Car dataset and Indoor dataset.}} }
\label{tab:baselines}
\end{center}
\vspace{-0.5cm}
\end{table}

\begin{figure}
  \center
    \begin{minipage}[t]{0.0465\linewidth}
    \includegraphics[width=\linewidth]{gfx/trans_x.pdf}
  \end{minipage}\begin{minipage}[t]{0.953\linewidth}
         \includegraphics[width=\linewidth]{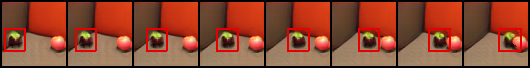}
  \end{minipage}
    \begin{minipage}[t]{0.0465\linewidth}
    \includegraphics[width=\linewidth]{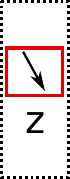}
  \end{minipage}\begin{minipage}[t]{0.953\linewidth}
         \includegraphics[width=\linewidth]{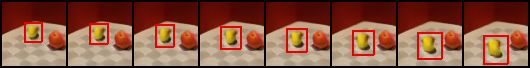}
  \end{minipage}
    \caption{{\bf Fruit Dataset.} We translate one fruit in each row.} 
    \label{fig:fruit}
    \vspace{-0.3cm}
\end{figure}

\boldparagraph{Real Data}
Qualitative results of our method on the real-world dataset are shown in \figref{fig:fruit}. 
We see that our method is able to synthesize plausible images even from real data.
Making use of our disentangled 3D representation, we are able to change the arrangement of fruits while our model properly handles occlusions and shadows.

\subsection{Ablation Study}

We now compare different representations and analyze multiple variants of our framework in an ablation study.
As we focus on the foreground objects in this study, we use the Car dataset without background for all experiments.

\begin{figure}
  \begin{center}
  \def\mywidth{0.1}
  \setlength\tabcolsep{9.pt}
  \begin{tabular}{c@{\hskip 2pt}c@{\hskip 2pt}c@{\hskip 20pt}c@{\hskip 20pt}c} 
    \includegraphics[width=\mywidth\linewidth]{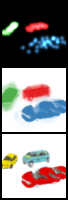} &
    \includegraphics[width=\mywidth\linewidth]{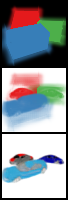} &
    \includegraphics[width=\mywidth\linewidth]{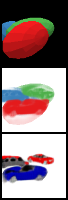} &
    \includegraphics[width=\mywidth\linewidth]{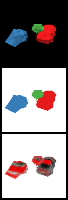} & 
    \includegraphics[width=\mywidth\linewidth]{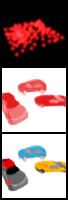} \\
  (a)&(b)&(c)&(d)&(e)\\
  \end{tabular}
  \vspace{-0.75em}
  \caption{{\bf Ablation Study.} (a) point cloud, (b) cuboid, (c) sphere, (d) deformable sphere w/o  $g^{2D}_\theta$, (e) single primitive. The first two rows show the projected primitives and the composite $\bA'_i$ (colors denote instances). Note that we visualize all primitives in one image while during training they are rendered and fed into the 2D generator individually.}
  \label{fig:representation}  
  \end{center}
  \vspace{-0.5cm}
\end{figure}

\boldparagraphwocolon{Point cloud, sphere or cuboid?}
We first compare the performance of different 3D representations for the task of 3D controllable image synthesis. \tabref{tab:representation} compares different representations in terms of their 3D controllability (FID$_\bt$ and FID$_\bR$) and disentanglement capability (FID$_i$).
We observe that different representations achieve very similar FID 
scores,
suggesting that the precise form of the 3D representation is less relevant than the general concept of a joint 3D-2D representation.
This can also be observed from \figref{fig:representation} which illustrates the different representations and the corresponding generated images.
The sphere representation performs best overall. %
We hypothesize that it is superior to the cuboid representation as it does not suffer as much from surface discontinuities in 3D space.

\begin{figure}
  \setlength\tabcolsep{0.2em}
  \begin{tabular}{m{0.5em}m{22.5em}}
    {\rotatebox[origin=c]{90}{\footnotesize{Single.}}} &
    \includegraphics[width=\linewidth]{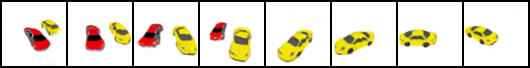}\\
    {\rotatebox[origin=c]{90}{\footnotesize{w/o $\cL_{geo}$}}} &
    \includegraphics[width=\linewidth]{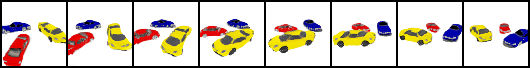}\\
    {\rotatebox[origin=c]{90}{\footnotesize{$\cL_{geo}$}}} &
    \includegraphics[width=\linewidth]{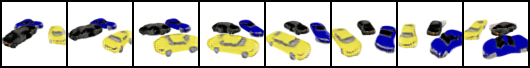}
  \end{tabular}
    \caption{{\bf Evaluation of Rotation Consistency.} Images sampled from $180^{\circ}$ of camera rotation. Top row: single primitive representation. The number of objects varies wrt. camera rotation. Middle and bottom rows: multi-primitive model w/o and with the geometric consistency loss. } 
    \label{fig:3d_consistency_loss}
    \vspace{-0.3cm}
\end{figure}

\boldparagraphwocolon{Can we learn the 3D model directly?} 
Inspired by previous works~\cite{Kato2018CVPR,Liu2019soft} that learn a 3D mesh directly from 2D images in the discriminative model, we also explore
whether we can directly learn an accurate 3D model in our generative setting.
In other words, we investigate if the 2D generator is helpful or not.
To answer this question, we remove the 2D generator from the pipeline and directly add the adversarial loss on the rendered image $\bX_i$.
This tasks the generative 3D model to predict accurate and textured 3D reconstructions.
To allow the generator to modify the shape, we learn to deform a sphere similar to recent mesh-based reconstruction techniques~\cite{Wang2018ECCV}.
\figref{fig:representation}~(d) shows that it is indeed possible to learn a reasonable shape this way. However, the image fidelity is inferior compared to images generated from our 2D generator as shown in \tabref{tab:representation}.

\boldparagraphwocolon{What is the advantage of disentanglement?}
To investigate the advantage of our object-wise generation pipeline, we create a variant of our method using only a single primitive to represent the entire scene. This is illustrated in \figref{fig:representation}~(e) and evaluated quantitatively in \tabref{tab:representation}. More specifically, we render a point cloud with $N\times M$ points to ensure the model has the same capacity as our multi-primitive model. We do not evaluate FID$_i$ as the model is not able to generate single object images. Not surprisingly, the single-primitive model is also able to generate images with low FID scores. However, this technique does not allow full control over the individual 3D objects. Moreover, the number of objects changes when camera viewpoint changes as shown in \figref{fig:3d_consistency_loss} (top row). 
\begin{table}[t]
\begin{center}
\begin{small}

\begin{tabular}{p{0.24\textwidth}p{0.025\textwidth}p{0.025\textwidth}p{0.025\textwidth}p{0.025\textwidth}p{0.025\textwidth}}
\toprule
{}   &     FID &   FID$_\bt$ &   FID$_\bR$ &   FID$_i$ \\
\hline
Vanilla GAN~\cite{Mescheder2018ICML}			&  50 & -- & -- & 41   \\
\hline
Point cloud 	&           38 &           43 &        44 &           66 \\
Cuboid		 &           38 &           45 &           45 &           60 \\
Sphere  	 &           33 &           45 &           45 &           53 \\
\hline
Deformable primitive w/o $g^{2D}_\theta$   &           69 &           71 &           74 &           69  \\
Single primitive 		&  30 &   38 &  44 &           -- \\
\bottomrule
\end{tabular}

\end{small}
\caption{{{\bf Ablation Study.} FID on Car dataset without background wrt. different primitive representations and architecture components.} }
\label{tab:representation}
\end{center}
\vspace{-0.8em}
\end{table}

\boldparagraphwocolon{Is the learned representation consistent in 3D?}
We aim for a model that can generate consistent scenes across camera viewpoints and 3D object poses. To test this ability, we rotate the camera and generate new samples as shown in \figref{fig:3d_consistency_loss}.
Our method generates locally coherent images due to its architecture design. In contrast, the single primitive baseline fails to preserve the object identity. In addition, we evaluate if adding the geometric consistency loss can further improve performance in \figref{fig:3d_consistency_loss}.
While we do not observe significant performance gains, geometric consistency cues allow for learning to synthesize depth alongside object appearance. See supplementary for qualitative results.

\subsection{Failure Cases}
\begin{figure}
  \center
  \setlength\tabcolsep{0em}
  \begin{tabular}{m{1.05em}m{22.5em}}
    \includegraphics[width=\linewidth]{gfx/trans_x.pdf} &
    \includegraphics[width=\linewidth]{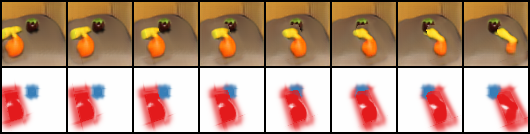}\\
    \includegraphics[width=\linewidth]{gfx/cam_rot_azimuth.pdf} &
    \includegraphics[width=\linewidth]{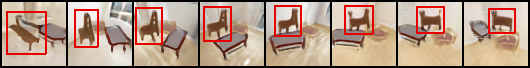}\\
  \end{tabular}
  \caption{{\bf Failure Cases.} Two fruits are generated from the same primitive (red in the alpha map) in the first example. The object identity changes in the second example.
  }
  \label{fig:failure}
  \vspace{-0.5cm}
\end{figure}

We show several failure cases of our method in \figref{fig:failure}.
Occasionally, a single primitive generates multiple objects (top)
or the object's identity changes when the viewpoint change is very large (bottom).
We believe that stronger inductive biases are required to tackle these problems.

\section{Conclusion}\label{sec:conclution}

We consider this paper as a first step towards unsupervised learning of 3D controllable image synthesis. 
We demonstrate that modeling both in 3D and 2D space is crucial for accurate and view-consistent results.
In addition, we show that our method is able to successfully disentangle scenes with multiple objects while providing controllability in terms of camera viewpoint and object poses.
We believe that incorporating stronger inductive biases about object shape or scene geometry will be key for tackling more challenging scenarios.
We hope that our results on this new task inspire further research in this exciting area.

\section*{Acknowledgments}
We acknowledge support from the BMBF through the Tübingen AI Center (FKZ: 01IS18039B).

{\small
	\bibliographystyle{ieee}
	\bibliography{bibliography_long,bibliography,bibliography_custom}
}

\end{document}